\documentclass{article}

\PassOptionsToPackage{numbers, compress}{natbib} % add
 \usepackage[preprint]{neurips_2026}

\usepackage[utf8]{inputenc} % allow utf-8 input
\usepackage[T1]{fontenc}    % use 8-bit T1 fonts
\usepackage{url}            % simple URL typesetting
\usepackage{booktabs}       % professional-quality tables
\usepackage{amsfonts}       % blackboard math symbols
\usepackage{nicefrac}       % compact symbols for 1/2, etc.
\usepackage{microtype}      % microtypography
\usepackage{xcolor}         % colors
\usepackage[colorlinks,
            linkcolor=red,
            anchorcolor=blue,
            citecolor=green
            ]{hyperref}
\usepackage{caption}
\usepackage{subcaption}
\usepackage{natbib}
\usepackage{graphicx}
\usepackage{enumitem}
\usepackage{makecell}
\usepackage{amsmath}
\usepackage{amsthm}
\usepackage{thmtools}
\usepackage{thm-restate}
\usepackage{algorithm}
\usepackage{algorithmic}
\usepackage{bbm}
\usepackage{bm}
\usepackage{bbding} 
\usepackage{multirow}

\usepackage{eso-pic}
\newcommand{\longcatlogoposfirst}{\AtPageUpperLeft{\hspace{37.7mm}\raisebox{-26.3mm}{\includegraphics[height=9mm]{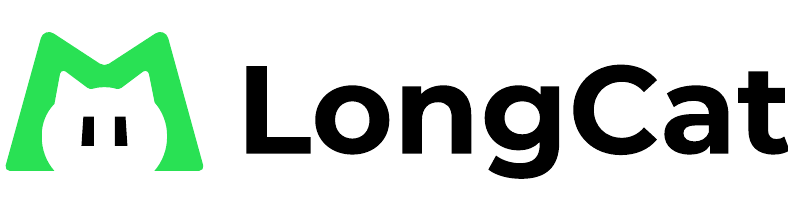}}}}

\AddToShipoutPictureFG{%
  \ifnum\value{page}=1\relax
    \longcatlogoposfirst
  \fi
}

\usepackage{xspace}
\newcommand{\ours}{NoisyAgent\xspace}

\definecolor{-}{rgb}{0.25,0.41,0.88}
\definecolor{+}{rgb}{0.70,0.13,0.13}

\title{Learning to Act under Noise:\\ Enhancing Agent Robustness via Noisy Environments}

\author{
  \textbf{Yuxin Chen$^{1,2,*}$},
  \textbf{Xiaodong Cai$^{2,3,*}$},
  \textbf{Junfeng Fang$^{1}$},
  \textbf{Zhuowen Han$^{2,4}$},
  \\
  \textbf{Yu Wang$^{2,5}$},
  \textbf{Yaorui Shi$^{2,5}$},
  \textbf{Yi Zhang$^{2,5}$},
  \textbf{Qi Gu$^{2,\dagger}$},
  \\
  \textbf{Xunliang Cai$^{2}$},
  \textbf{Xiang Wang$^{5}$},
  \textbf{An Zhang$^{5,\dagger}$},
  \textbf{Tat-Seng Chua$^{1}$}
  \\
  \vspace{-2mm} \\
  $^{1}$National University of Singapore,
  $^{2}$Meituan, \\
  $^{3}$Tsinghua University,
  $^{4}$Tianjin University, \\
  $^{5}$University of Science and Technology of China \\
  $^{*}$Equal contribution. \\
  $^{\dagger}$Corresponding authors: \texttt{guqi03@meituan.com}, \texttt{an\_zhang@ustc.edu.cn}
}

\begin{document}

\maketitle

\begin{abstract}
\label{abstract}

Recent advances in large language models (LLMs) have facilitated the widespread deployment of LLMs as interactive agents capable of reasoning, planning, and tool use.
Despite strong performance on existing benchmarks, such agents often exhibit notable degradation when deployed in real-world settings, where environments are inherently stochastic and imperfect. 
We argue that this discrepancy arises from a fundamental mismatch between idealized training settings and real-world interaction dynamics, where current paradigms rely on carefully curated task instructions and stable, well-controlled environments.
To address this gap, we propose \ours, an agentic training framework that explicitly incorporates environmental imperfections into the agent learning process. 
We identify two major sources of interaction noise in real-world scenarios: user noise, which captures ambiguity and variability in user interaction, and tool noise, which reflects failures and anomalies in tool execution. 
We introduce such perturbations into the training pipeline by modifying user interaction patterns and simulating tool execution results within the training environment. 
To stabilize training while encouraging agents to handle increasingly challenging imperfection, noise is applied to only a subset of rollouts and progressively increased in difficulty as the model adapts to the current noise level.
Extensive experiments demonstrate that our approach consistently improves agent robustness under noisy and dynamic environments. 
Our analysis reveals that training under noise condition also yields performance gains on idealized benchmarks, suggesting that controlled exposure to environmental noise promotes more generalizable reasoning and decision-making behaviors. 
Our findings highlight the importance of modeling interaction imperfections for bridging the gap between agent training and real-world deployment.

\end{abstract}
% \newpage
\section{Introduction}
\label{introduction}

Recent advances in large language models (LLMs) have transformed them from passive text generators into interactive agents capable of reasoning, planning, and tool use~\citep{gpt52,gemini3pro,team2025longcat-flash-thinking}, enabling their widespread deployment in real-world applications. 
As these capabilities continue to improve~\citep{team2025kimi,zeng2025glm,team2025longcat}, LLM agents have achieved strong performance across a wide range of benchmarks~\citep{yao2024tau,barres2025tau,he2025vitabench}. 
However, this success does not consistently transfer to more realistic settings: when confronted with complex and dynamic environments, many agents exhibit notable performance degradation~\citep{deng2023mind2web, zhou2023webarena, xue2025illusion}.

We argue that current agent learning paradigms exhibit a fundamental gap between training conditions and real-world deployment.
A common characteristic shared by existing agent training paradigms is their reliance on idealized assumptions, where agents are trained with carefully curated instructions and interact with stable, well-controlled environments~\citep{zeng2024agenttuning, qi2024webrl, jin2025search}. 
In contrast, real-world environments are inherently stochastic and imperfect. 
Users often exhibit diverse interaction styles and unpredictable behaviors~\citep{gallois2005communication,trippas2024users,wang2024understanding}, while external tools may return noisy, incomplete, or even failed outputs due to various uncontrollable factors~\citep{vuddanti2025paladin, xiong2025butterfly}. 
This discrepancy between training conditions and deployment environments limits the robustness of current agents, often leading to degraded performance in practical applications~\citep{rawles2024androidworld, sun2025gui, shoeb2025out}.

\begin{figure}[t]
    \centering
    % 0.8\linewidth
    \includegraphics[width=0.98\linewidth]{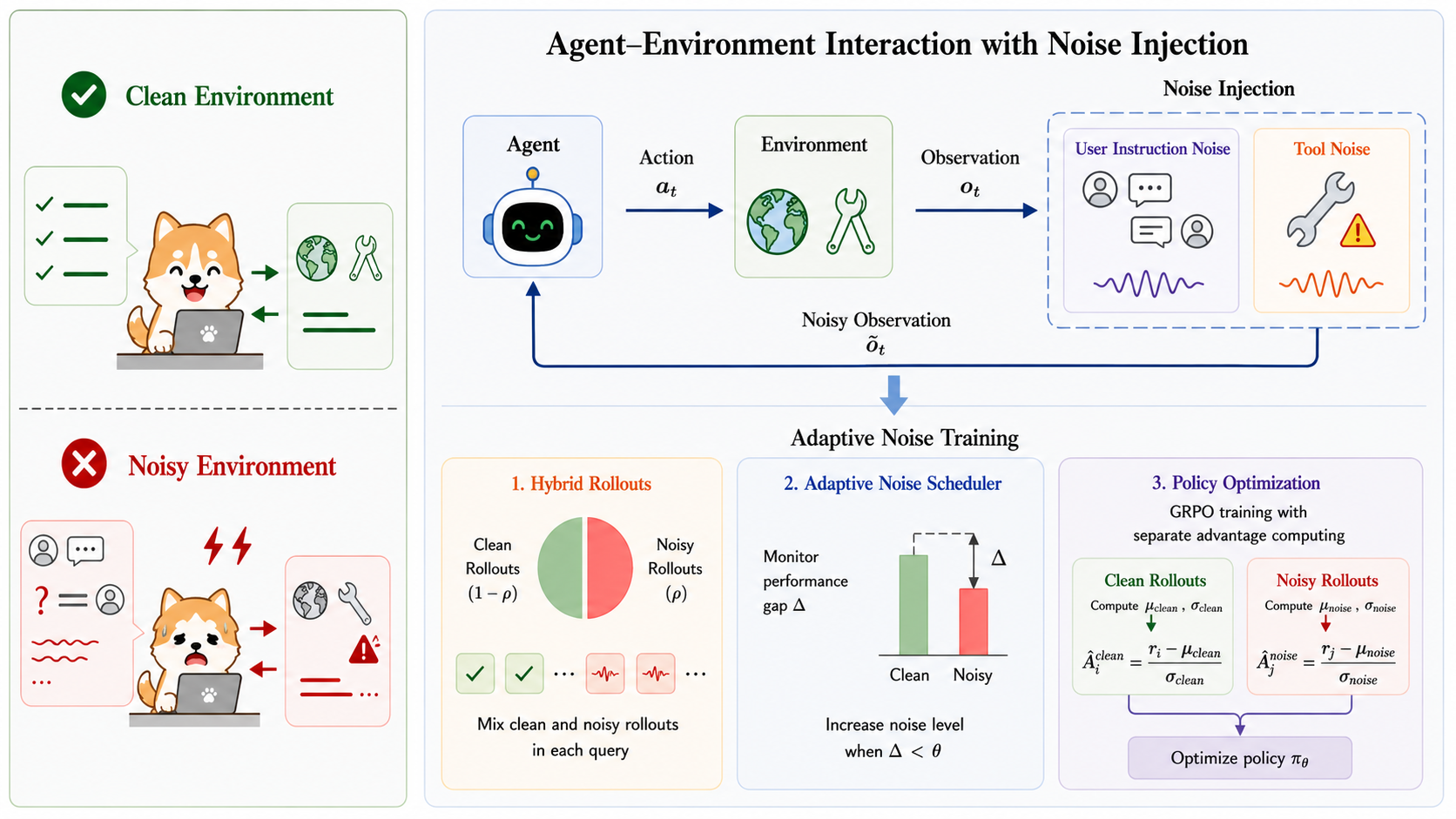}
    \caption{
    Overview of \ours. 
    We inject structured perturbations into both user instructions and tool responses to simulate real-world imperfections. 
    Training is conducted via hybrid rollouts that combine clean and noisy trajectories, together with an adaptive scheduler that increases noise difficulty based on the performance gap $\Delta$. 
    Policy optimization is performed with group-wise normalization to stabilize learning under heterogeneous interaction conditions.
    }
    \label{fig:method}
\end{figure}

Inspired by the success of stochastic perturbations in reinforcement learning~\citep{tobin2017domain, sadeghi2017cad2rl, zhao2021robust}, we argue that agent robustness emerges from exposure to diverse imperfections in learning process.
Rather than relying on idealized training settings and expecting agents to adapt post hoc, we explicitly incorporate environmental noise and uncertainty into the agentic training process. 
However, how to model and introduce such noise in agentic training remains underexplored, and naively injecting noise into the training environment can easily destabilize training dynamics, making it a non-trivial challenge.

Toward this goal, we propose \ours, an agentic RL method for training under noisy environments.
We begin by identifying representative forms of real-world noise and developing an automated pipeline to incorporate such imperfections into the training process.
Concretely, we consider two major sources of interaction noise in real-world agent scenarios: user noise, which captures ambiguity and variability in user interactions, and tool noise, which simulates execution anomalies from external tools.
These perturbations are introduced by modifying user instructions and simulating tool execution results within the training environment, with perturbations applied to only a subset of rollouts for each task.
Training follows a curriculum schedule. Starting from mild perturbations, we progressively increase the difficulty and ratio of noise as the model exhibits sufficient robustness at each stage.
Robustness is quantified by the performance gap between idealized and perturbed environments on the same tasks.
This adaptive process ensures that training remains informative rather than overwhelming, while avoiding inefficient exploration of excessively noisy regimes.

Benefiting from our noise-aware training, agents achieve improved performance on benchmarks augmented with real-world noise, indicating enhanced robustness under imperfect and dynamic environments.
Interestingly, we also observe consistent gains on standard, idealized benchmarks. 
We hypothesize that appropriately designed noise introduces controlled instability into the training environment and promotes more generalizable reasoning and decision-making. In particular, exposure to noisy and uncertain interactions encourages agents to recover from errors, resolve ambiguities, and adapt to unexpected outcomes. From this perspective, noise serves as a form of implicit difficulty augmentation, enriching the training distribution and improving robustness beyond idealized settings.

Overall, our contributions can be concluded as follows:
\begin{itemize}[leftmargin=*]
    \item We identify a fundamental gap between idealized agent training and real-world deployment, highlighting the importance of modeling environmental uncertainty for robust agent learning.
    \item We develop a noise-aware training framework that systematically incorporates instruction and tool perturbations into the training environment.
    \item Extensive experiments demonstrate that our approach consistently improves agent robustness under noisy and dynamic environments, while also yielding performance gains on standard benchmarks.
\end{itemize}

\section{Preliminary}
\label{sec:preliminary}

\subsection{Agentic Reinforcement Learning}

In representative agentic training paradigm, each taks can be formalized as a Partially Observable Markov Decision Process (POMDP)~\citep{agentrl_survey2025}:
\begin{equation}
\mathcal{M} = (\mathcal{S}, \mathcal{A}, \mathcal{O}, \mathcal{T}, \mathcal{R}). 
\end{equation}
At each step $t$, the agent maintains a state $s_t = (s_t^{\text{env}}, h_t, q) \in \mathcal{S}$, which captures the environment state $s_t^{\text{env}}$, the interaction history $h_t$, and the task prompt $q$. 
Based on the current observation $o_t \in \mathcal{O}$, the agent selects an action $a_t \in \mathcal{A}$, where the action space $\mathcal{A} = \mathcal{A}_{\text{user}} \cup \mathcal{A}_{\text{tool}}$ includes both user interaction and tool calling invocations. 
Correspondingly, the observation space $\mathcal{O} = \mathcal{O}_{\text{user}} \cup \mathcal{O}_{\text{tool}}$ consists of user-side feedback and tool execution results. 
Upon taking action $a_t$, the environment states evolves according to the transition function $\mathcal{T}: \mathcal{S} \times \mathcal{A} \rightarrow \mathcal{S} \times \mathcal{O}$, producing the next observation $o_{t+1}$. 
The training objective is to learn a policy $\pi_\theta$ that maximizes the expected cumulative reward $\mathbb{E}_{\tau \sim \pi_\theta}\!\left[\sum_{t=0}^{T} r_t\right]$ over trajectories $\tau = (o_0, a_0, o_1, a_1, \ldots, o_T)$.

A widely adopted training paradigm is Reinforcement Learning with Verifiable Rewards (RLVR)~\citep{deepseek_r1,verltool2025}, where a verifier evaluates whether the final environment state $s_T^{\text{env}}$ or the full trajectory $\tau$ satisfies the task instruction given rubrics, providing a scalar reward at the trajectory level.
To optimize the policy, a representative approach is Group Relative Policy Optimization (GRPO)~\citep{shao2024grpo}, which extends PPO~\cite{schulman2017ppo} by computing advantages relative to a group of sampled rollouts.
Concretely, given a task prompt $q$ and $G$ sampled trajectories $\{\tau_1, \ldots, \tau_G\}$, the advantage of each trajectory is computed as $\hat{A}_i = (r_i - \mu)/\sigma$, where $\mu$ and $\sigma$ are the mean and standard deviation of the group rewards. 
The objective can be written as:
\begin{equation}
\mathcal{J}_{\text{GRPO}}(\theta) = \mathbb{E}_{q}\left[\frac{1}{G}\sum_{i=1}^{G} \frac{1}{L_i}\sum_{t=1}^{L_i}
\min\!\left(\rho_{i,t}\,\hat{A}_i,\; \text{clip}(\rho_{i,t},\, 1\!\pm\!\epsilon)\,\hat{A}_i\right)
\right].
\end{equation}
where $\rho_{i,t} = \frac{\pi_\theta(a_{i,t}\mid h_{i,t})}{\pi_{\text{old}}(a_{i,t}\mid h_{i,t})}$ and $L_i$ is the length of trajectory $\tau_i$. 
Building on this standard optimization paradigm, effective agentic training relies on access to a diverse set of interactive environments that support both user-agent interaction and tool-grounded execution~\cite{longcatflash2601,liu2025deepseekv32}. 
% However, constructing such environments at scale remains a significant challenge.

\subsection{Scaling Environment for Agentic Training}

Constructing interactive environments manually for agentic training is costly and difficult to scale. 
Recent work addresses this challenge by synthesizing executable environments from high-level domain specifications in a fully automated environment scaling pipeline~\citep{scaleenv2025}. 
Given a domain definition, the pipeline initializes a domain-specific tool set together with a unified database schema, forming a structured domain graph $\mathcal{D}$ that serves as the foundation for executable environment generation. 
By sampling from this graph, each training environment can be instantiated as consisting of two tightly coupled components: a user-side construction that specifies task objectives and interaction patterns, and a tool-side construction that defines environment dynamics.

On the user side, tasks are synthesized by sampling tool chains from the domain graph and generating corresponding task queries together with interaction patterns, resulting in compositional objectives that specify both what to solve and how the user agent interacts within the environment. 
Formally, the user-side construction can be expressed as:
\begin{equation}
(q, u_{\text{int}}) = f_{\text{user}}(\mathcal{D}),
\end{equation}
where $q$ is the task prompt and $\pi_{\text{int}}$ denotes the interaction pattern governing user-agent interactions.
$f_{\text{user}}$ denote simplified abstractions of user-side construction processes.

On the tool side, complete executable environments are constructed by implementing structured tool APIs and underlying environment databases based on the domain graph. 
The sampled tool chains are instantiated as reference executions, and the tool set is further expanded along the domain graph while ensuring both correctness and verifiability of the execution process. 
Formally, the tool-side construction can be written as:
\begin{equation}
\mathcal{E} = f_{\text{tool}}(\mathcal{D}, q, u_{\text{int}}),
\end{equation}
where $\mathcal{E}$ defines the executable environment grounded in the task specification, including tool APIs, valid state transitions, and verifiable execution paths. $f_{\text{tool}}$ denote simplified abstractions of tool-side construction processes.

While this design enables scalable and reliable task construction, it assumes that both components are well-specified: user interactions are restricted to be clear and helpful, while tool behaviors are stable. 
As a result, the resulting training environments are often idealized, leading to a mismatch between training and deployment, where real-world environments are inherently imperfect.
\section{Methodology} 
\label{sec:methodology}

To bridge the gap between idealized training and noisy deployment, we propose \ours, an agentic training framework that explicitly incorporates environmental imperfections into learning. 
We first introduce an automatic noise injection pipeline (Section~\ref{subsec:noise injection}) that augments training with user- and tool-side perturbations, and then present an adaptive training strategy (Section~\ref{subsec:noise training}) that progressively adjusts noise difficulty to ensure stable and effective learning.

\subsection{Automatic Noise Injection}
\label{subsec:noise injection}

We systematically analyze common real-world noise and design an automated pipeline to explicitly incorporate these imperfections into any synthesized agentic training environment. 
Concretely, we consider two major sources of interaction noise in real-world agentic scenarios: user-side noise, which captures ambiguity and variability in user interaction patterns, and tool-side noise, which reflects failures and anomalies in external tool execution.
To model such imperfections, we introduce a noise generator $\pi_{\text{noise}}$ that stochastically perturbs the agent–environment interaction at each step to simulate imperfect observation from the real world.

\paragraph{User-side Injection.}
On the user side, noise is injected before the task starts by modifying the interaction patterns specified by the user.
We simulate representative non-ideal interaction patterns observed in real-world scenarios, including: 
(1) \textit{Ambiguous}, where user intent is underspecified; 
(2) \textit{Inconsistent}, where user needs change or conflict over time; and
(3) \textit{Redundant}, where irrelevant or unnecessary information is included.
Formally, given the interaction pattern $u_{\text{int}}$ defined by any environment scaling pipeline, the injection of user-side noise can be expressed as:
\begin{equation}
\tilde{u}_{\text{int}} = \pi_{\text{noise}}(u_{\text{int}}),
\end{equation}
where $\tilde{u}_{\text{int}}$ denotes the perturbed counterpart. 
This transformation introduces additional variability and ambiguity into user–agent interactions. 
To avoid inducing unreliable or misleading reward signals, we preserve the underlying task objective $q$, ensuring that the injected perturbations do not invalidate task solvability, but instead increase the difficulty and stochasticity of the interaction process.

\paragraph{Tool-side Injection.}
Tool-side noise is injected during agent rollouts by randomly perturbing a subset of tool execution results to simulate stochasticity in real-world environments. 
Specifically, we model common execution anomalies in real-world systems, including: 
(1) \textit{Failures}, where tool requests return errors; 
(2) \textit{Incomplete}, where outputs are truncated; 
(3) \textit{Misleading}, where responses contain incorrect or inconsistent information; and 
(4) \textit{Redundant}, where outputs include unnecessary details. 
Formally, the injection of tool-side noise can be formulated as:
\begin{equation}
\tilde{o}_{t} = \pi_{\text{noise}}(o_{t}),
\end{equation}
where $o_t$ denotes the original tool response and $\tilde{o}_t$ is the perturbed output. 
This process simulates imperfect tool behaviors while maintaining executable interaction dynamics.

\subsection{Adaptive Noise Training}
\label{subsec:noise training}

\paragraph{Hybrid Training.}
The proposed automatic noise injection pipeline enables the incorporation of imperfections into agent training process. 
However, agent learning is highly sensitive to both task instructions and environment feedback, naively injecting uncontrolled noise can destabilize training dynamics. 
To preserve training stability while improving robustness, we adopt a hybrid training scheme that combines idealized and perturbed environments.

Concretely, under the GRPO training paradigm, given a task set $\mathcal{Q}$, we sample a task $q \in \mathcal{Q}$ and perform $N$ independent rollouts in parallel environments. 
Among these, a subset of $N_{\text{noise}}$ rollouts are perturbed by injecting user-side or tool-side noise with a controllable difficulty level, while the remaining $N - N_{\text{noise}}$ rollouts are conducted in clean, idealized environments.

Formally, let $\mathcal{T}_{\text{clean}}$ and $\mathcal{T}_{\text{noise}}$ denote the sets of clean and perturbed trajectories for a given task $q$, respectively. 
In our setting, rollouts are partitioned into these two groups, and we modify the standard GRPO objective by computing advantages separately within each group while optimizing over their union. 
The overall objective is defined as:
\begin{equation}
\mathcal{J}(\theta) = 
\mathbb{E}_{q}\left[
\frac{1}{G} \left(
\sum_{i \in \mathcal{T}_{\text{clean}}} \frac{1}{L_i} \sum_{t=1}^{L_i}
\mathcal{L}_{i,t}(\hat{A}_i^{\text{clean}})
+
\sum_{j \in \mathcal{T}_{\text{noise}}} \frac{1}{L_j} \sum_{t=1}^{L_j}
\mathcal{L}_{j,t}(\hat{A}_j^{\text{noise}})
\right)
\right],
\end{equation}
where
\begin{equation}
\mathcal{L}_{k,t}(\hat{A}) =
\min\!\left(
\rho_{k,t} \hat{A},\;
\text{clip}(\rho_{k,t}, 1 \pm \epsilon)\hat{A}
\right), \quad
\rho_{k,t} = \frac{\pi_\theta(a_{k,t}\mid h_{k,t})}{\pi_{\text{old}}(a_{k,t}\mid h_{k,t})}.
\end{equation}

The advantages are computed separately within each group:
\begin{equation}
\hat{A}_i^{\text{clean}} = \frac{r_i - \mu_{\text{clean}}}{\sigma_{\text{clean}}}, \quad
\hat{A}_j^{\text{noise}} = \frac{r_j - \mu_{\text{noise}}}{\sigma_{\text{noise}}},
\end{equation}
where $\mu_{\text{clean}}, \sigma_{\text{clean}}$ and $\mu_{\text{noise}}, \sigma_{\text{noise}}$ denote the mean and standard deviation of rewards computed within each group. 
This group-wise normalization prevents the dominance of either clean or noisy rollouts during optimization, and stabilizes training under heterogeneous interaction conditions. 

\paragraph{Noise Scheduling.}

To adaptively introduce noise while maintaining training stability, we first quantify the model's robustness to different noise types and adjust the noise level accordingly.

We measure the model's robustness to a specific noise type via the performance gap between clean and perturbed rollouts on the same task:
\begin{equation}
\Delta = \mathbb{E}_{\tau \sim \mathcal{T}_{\text{clean}}}[\mathbf{1}(r(\tau)=1)] 
- \mathbb{E}_{\tau \sim \mathcal{T}_{\text{noise}}}[\mathbf{1}(r(\tau)=1)],
\end{equation}
where $r(\tau)=1$ indicates successful task completion. 
This gap reflects the extent to which current noise degrades task performance.

Based on this measure, we adopt a progressive noise scheduling strategy. 
Training is initialized in fully idealized environments, with noise gradually introduced as the model adapts. 
At each stage, we control two factors: (i) the noise scale, defined as the proportion of perturbed rollouts $\rho = N_{\text{noise}} / N$; and (ii) the noise difficulty, characterized by the frequency of tool-side perturbations and the severity of user-side interaction anomalies.
When $\Delta < \theta$, with $\theta$ denoting a predefined threshold, the model is considered to have adapted to the current noise level, and we increase both the difficulty and the proportion of that noise type. 
This yields a curriculum over noise, progressively increasing interaction complexity while maintaining training stability.
\section{Experiments} 
\label{sec:experiments}

\subsection{Experiment Settings}

\paragraph{Training Environment.}
Our training environment follows the environment scaling pipeline of~\cite{scaleenv2025}. 
Within the synthesis pipeline, we leverage a diverse suite of high-performance LLMs for different roles. 
Specifically, GPT-4.1 is used for environment construction due to its favorable trade-off between cost and efficiency, while Claude-Sonnet-4.5 serves as a verifier given its strong evaluation capability. 
GLM-4.6 is employed to synthesize diverse instructions, forming the basis of our RL task set. 
Building on the synthesized tasks, we use Qwen2.5-72B-Instruct as a noise injector to introduce controlled perturbations into the interaction process. 
During training, Qwen2.5-72B-Instruct also acts as the user simulator to generate natural language feedback, while a Qwen3-32B model is trained as an evaluator to assign rewards based on the synthesized rubrics.

\paragraph{Evaluation.}
We evaluate the robustness of the model on AgentNoiseBench~\cite{wang2026agentnoisebench}, a benchmark designed to assess agent performance under real-world noise.
We select two representative subsets, AgentNoiseBench-$\tau^2$ and AgentNoiseBench-Vita for evaluation. 
To assess performance in idealized environments, we evaluate on representative standard agent benchmarks: 
(i) $\tau^2$-Bench, a dual-control conversational benchmark where both the user and the agent can invoke tools in customer-service domains such as retail, airline, and telecom; 
(ii) Vita-Bench, a multi-tool agent benchmark covering real-world scenarios including food delivery, in-store services, and travel. 
Across all benchmarks, GPT-4.1 is used as the user simulator, and Claude-Sonnet-4.5 is used as the evaluator. 
Each experiment is repeated four times. We report Avg@4 and Pass@4 metrics averaged across tasks.

\paragraph{Implementation Details and Baselines.}
We adopt Qwen3-8B and Qwen3-32B as backbone models. 
On these backbones, we compare several representative training methods, including GRPO, DAPO, and GSPO, where our method is based on GSPO. 
The training batch size is set to 32, with 64 rollouts per sample. 
The proportion of noisy trajectories is capped at 50\% of the total rollouts. 
We set the scheduling threshold $\Delta$ to 0.05. 
The maximum prompt length is 8,192 tokens, and the maximum response length is 32,768 tokens. 
Detailed training configurations are provided in Appendix~\ref{appendix:training_config}.

\begin{table*}[t]
\centering
\scriptsize
\setlength{\tabcolsep}{3.5pt}
\caption{Main results under the noisy setting on AgentNoiseBench. We report Avg@4 and Pass@4 averaged across four runs. Best results are in bold, and second-best are underlined.}
\label{tab:main_results_noisy}
\begin{tabular}{c cc cc cc cc cc cc}
\toprule
\multirow{3}{*}{Method}
& \multicolumn{6}{c}{AgentNoiseBench-$\tau^2$}
& \multicolumn{6}{c}{AgentNoiseBench-Vita} \\
\cmidrule(lr){2-7} \cmidrule(lr){8-13}
&
\multicolumn{2}{c}{Retail}
& \multicolumn{2}{c}{Airline}
& \multicolumn{2}{c}{Telecom}
& \multicolumn{2}{c}{Delivery}
& \multicolumn{2}{c}{In-Store}
& \multicolumn{2}{c}{OTA} \\
\cmidrule(lr){2-3} \cmidrule(lr){4-5} \cmidrule(lr){6-7}
\cmidrule(lr){8-9} \cmidrule(lr){10-11} \cmidrule(lr){12-13}
&
Avg@4 & Pass@4
& Avg@4 & Pass@4
& Avg@4 & Pass@4
& Avg@4 & Pass@4
& Avg@4 & Pass@4
& Avg@4 & Pass@4 \\
\midrule

\textbf{Qwen3-8B}
& 24.12 & 44.74 & 23.00 & 42.00 & 21.05 & 41.23 & 11.75 & 18.00 & 8.50 & 12.00 & 0.75 & 2.00 \\
+ GRPO
& 30.48 & 50.88 & \underline{33.50} & \underline{54.00} & 31.58 & 53.51 & 15.25 & 24.00 & 14.25 & \underline{23.00} & 2.50 & 4.00 \\
+ DAPO
& 29.39 & 53.51 & 31.00 & 50.00 & 34.21 & \underline{57.89} & 15.75 & 25.00 & 12.75 & 19.00 & 2.25 & 4.00 \\
+ GSPO
& \underline{31.80} & \underline{54.39} & 32.50 & 52.00 & \underline{34.43} & 56.14 & \underline{16.00} & \underline{26.00} & \underline{15.00} & 22.00 & \underline{2.75} & \underline{5.00} \\
+ Ours
& \textbf{36.40} & \textbf{61.40} & \textbf{38.00} & \textbf{56.00} & \textbf{38.38} & \textbf{64.91} & \textbf{21.50} & \textbf{34.00} & \textbf{16.25} & \textbf{25.00} & \textbf{4.75} & \textbf{8.00} \\
\midrule

\textbf{Qwen3-32B}
& 31.14 & 52.63 & 31.50 & 56.00 & 26.54 & 45.61 & 19.50 & 30.00 & 14.75 & 21.00 & 5.50 & 9.00 \\
+ GRPO
& \underline{38.16} & \underline{61.40} & 37.00 & 62.00 & 36.84 & 62.28 & 23.25 & 35.00 & \underline{19.50} & \underline{28.00} & 7.25 & 11.00 \\
+ DAPO
& 36.18 & 57.89 & \underline{39.50} & \underline{66.00} & 38.16 & \underline{66.67} & \underline{24.00} & \underline{36.00} & 16.75 & 24.00 & \underline{7.50} & 11.00 \\
+ GSPO
& 37.72 & 60.53 & 39.00 & 64.00 & \underline{39.25} & 65.79 & 23.75 & \underline{36.00} & 17.50 & 25.00 & \underline{7.50} & \underline{12.00} \\
+ Ours
& \textbf{43.20} & \textbf{65.79} & \textbf{46.00} & \textbf{70.00} & \textbf{43.42} & \textbf{70.18} & \textbf{28.75} & \textbf{42.00} & \textbf{22.00} & \textbf{31.00} & \textbf{9.50} & \textbf{14.00} \\
\bottomrule
\end{tabular}
\end{table*}

\begin{table*}[t]
\centering
\scriptsize
\setlength{\tabcolsep}{3.5pt}
\caption{Main results under the ideal setting on standard agent benchmarks. We report Avg@4 and Pass@4 averaged across four runs. Best results are in bold, and second-best are underlined.}
\label{tab:main_results_ideal}
\begin{tabular}{c cc cc cc cc cc cc}
\toprule
\multirow{3}{*}{Method}
& \multicolumn{6}{c}{$\tau^2$-Bench}
& \multicolumn{6}{c}{VitaBench} \\
\cmidrule(lr){2-7} \cmidrule(lr){8-13}
&
\multicolumn{2}{c}{Retail}
& \multicolumn{2}{c}{Airline}
& \multicolumn{2}{c}{Telecom}
& \multicolumn{2}{c}{Delivery}
& \multicolumn{2}{c}{In-Store}
& \multicolumn{2}{c}{OTA} \\
\cmidrule(lr){2-3} \cmidrule(lr){4-5} \cmidrule(lr){6-7}
\cmidrule(lr){8-9} \cmidrule(lr){10-11} \cmidrule(lr){12-13}
&
Avg@4 & Pass@4
& Avg@4 & Pass@4
& Avg@4 & Pass@4
& Avg@4 & Pass@4
& Avg@4 & Pass@4
& Avg@4 & Pass@4 \\
\midrule

\textbf{Qwen3-8B}
& 35.31 & 59.65 & 27.00 & 52.00 & 22.59 & 42.98 & 13.75 & 22.00 & 15.50 & 24.00 & 1.75 & 4.00 \\
+ GRPO
& 46.05 & 73.68 & 36.50 & 62.00 & 37.28 & 57.89 & 21.00 & 33.00 & 22.75 & 35.00 & 4.25 & 7.00 \\
+ DAPO
& 44.52 & 71.05 & \underline{38.00} & \underline{66.00} & \underline{39.47} & \underline{63.16} & \underline{21.50} & \underline{34.00} & \underline{23.25} & \underline{36.00} & 4.00 & 7.00 \\
+ GSPO
& \underline{46.49} & \underline{74.56} & 37.50 & 64.00 & 39.04 & 61.40 & 21.25 & 33.00 & 23.00 & 35.00 & \underline{4.50} & \underline{8.00} \\
+ Ours
& \textbf{47.59} & \textbf{77.19} & \textbf{40.00} & \textbf{68.00} & \textbf{40.79} & \textbf{64.91} & \textbf{22.25} & \textbf{35.00} & \textbf{24.00} & \textbf{37.00} & \textbf{5.00} & \textbf{9.00} \\
\midrule

\textbf{Qwen3-32B}
& 49.12 & 72.81 & 38.00 & 66.00 & 28.95 & 49.12 & 23.00 & 35.00 & 26.00 & 38.00 & 7.00 & 12.00 \\
+ GRPO
& 58.11 & 83.33 & 45.00 & 72.00 & 41.67 & 68.42 & 27.00 & 40.00 & 30.25 & 43.00 & 8.75 & \underline{14.00} \\
+ DAPO
& 56.58 & 80.70 & \underline{47.50} & \underline{76.00} & 43.42 & \underline{71.93} & \underline{27.75} & \underline{41.00} & 29.50 & 42.00 & \underline{9.25} & \textbf{15.00} \\
+ GSPO
& \underline{58.55} & \underline{84.21} & 46.50 & 74.00 & \underline{43.86} & 70.18 & 27.25 & 40.00 & \underline{30.50} & \underline{44.00} & 9.00 & \underline{14.00} \\
+ Ours
& \textbf{60.31} & \textbf{86.84} & \textbf{49.50} & \textbf{78.00} & \textbf{45.39} & \textbf{78.07} & \textbf{29.00} & \textbf{43.00} & \textbf{32.25} & \textbf{46.00} & \textbf{9.75} & \textbf{15.00} \\
\bottomrule
\end{tabular}
\end{table*}

\subsection{Main Results}

Table~\ref{tab:main_results_noisy} and Table~\ref{tab:main_results_ideal} present the evaluation results under noisy and ideal settings, respectively. 
We have the following observations.

\paragraph{Noise-aware training significantly improves robustness under imperfect environments.}
Across all domains and both model scales, \ours consistently achieves the best performance on AgentNoiseBench, outperforming strong baselines such as GSPO and DAPO by a clear margin in both Avg@$4$ and Pass@$4$. 
In contrast, while standard RL methods improve performance under clean settings, their gains diminish substantially in the presence of noise, often exhibiting notable relative degradation across domains compared with their gains in idealized settings. 
This suggests that existing training paradigms are less effective when facing ambiguous user instructions and imperfect tool feedback. 
By incorporating structured perturbations during training, our method enables the agent to better handle uncertainty, recover from intermediate failures, and maintain consistent progress toward task completion under noisy conditions.

\paragraph{Training with noise leads to consistent gains even in idealized settings.}
Despite being designed for noisy environments, \ours also achieves consistent improvements on standard benchmarks without noise. 
Across both $\tau^2$-Bench and VitaBench, our method outperforms all baselines across domains and metrics. 
This indicates that training with noise does not harm performance in ideal settings, and can improve overall agent capability. 
We attribute this to the fact that exposure to diverse and imperfect interaction patterns encourages the agent to learn more robust and effective decision-making strategies, rather than relying on brittle interaction assumptions.

\subsection{Analysis}

\paragraph{Ablation Study.}

\begin{table}[t]
\centering
\small
\setlength{\tabcolsep}{4pt}
\caption{Ablation study of key components on Delivery domain of both AgentNoiseBench-Vita and VitaBench with Qwen3-8B. 
We report Avg@4 and Pass@4 averaged across four runs.}
\label{tab:ablation}
\begin{tabular}{lcccc}
\toprule
\multirow{2}{*}{Method} & \multicolumn{2}{c}{AgentNoiseBench-Vita} & \multicolumn{2}{c}{VitaBench} \\
\cmidrule(lr){2-3} \cmidrule(lr){4-5}
& Avg@4 & Pass@4 & Avg@4 & Pass@4 \\
\midrule
Ours & \textbf{21.50} & \textbf{34.00} & \textbf{22.25} & \textbf{35.00}  \\
w/o controlled injection & 13.25 & 21.00 & 14.75 & 24.00 \\
w/o scheduling & 20.00 & 31.00 & 21.50 & 33.00 \\
w/o noise & 16.00 & 26.00 & 21.25 & 33.00 \\
w/o training & 11.75 & 18.00 & 13.75 & 22.00 \\
\bottomrule
\end{tabular}
\end{table}

To isolate the effect of each component, we perform ablations by removing individual elements from our framework.
\emph{w/o controlled injection} removes the hybrid training scheme, applying noise to all rollouts instead of mixing clean and noisy trajectories. 
\emph{w/o scheduling} removes the curriculum over noise training, using perturbations of fixed complexity throughout training. 
\emph{w/o noise} reduces training to an idealized setting without any perturbations. 
\emph{w/o training} evaluates the base model without RL optimization.
Overall, removing any component leads to performance degradation, indicating that each part contributes to the final performance. 
In particular, uncontrolled noise injection (\emph{w/o controlled injection}) causes the largest drop, suggesting that naively introducing perturbations can destabilize training. 
In contrast, incorporating a proper scheduling strategy further improves performance, showing that progressively adjusting noise leads to more effective and stable learning.

\paragraph{Training Dynamics.}

\begin{figure}[t]
\centering
\begin{subfigure}[t]{0.45\columnwidth}
    \centering
    \includegraphics[width=\linewidth]{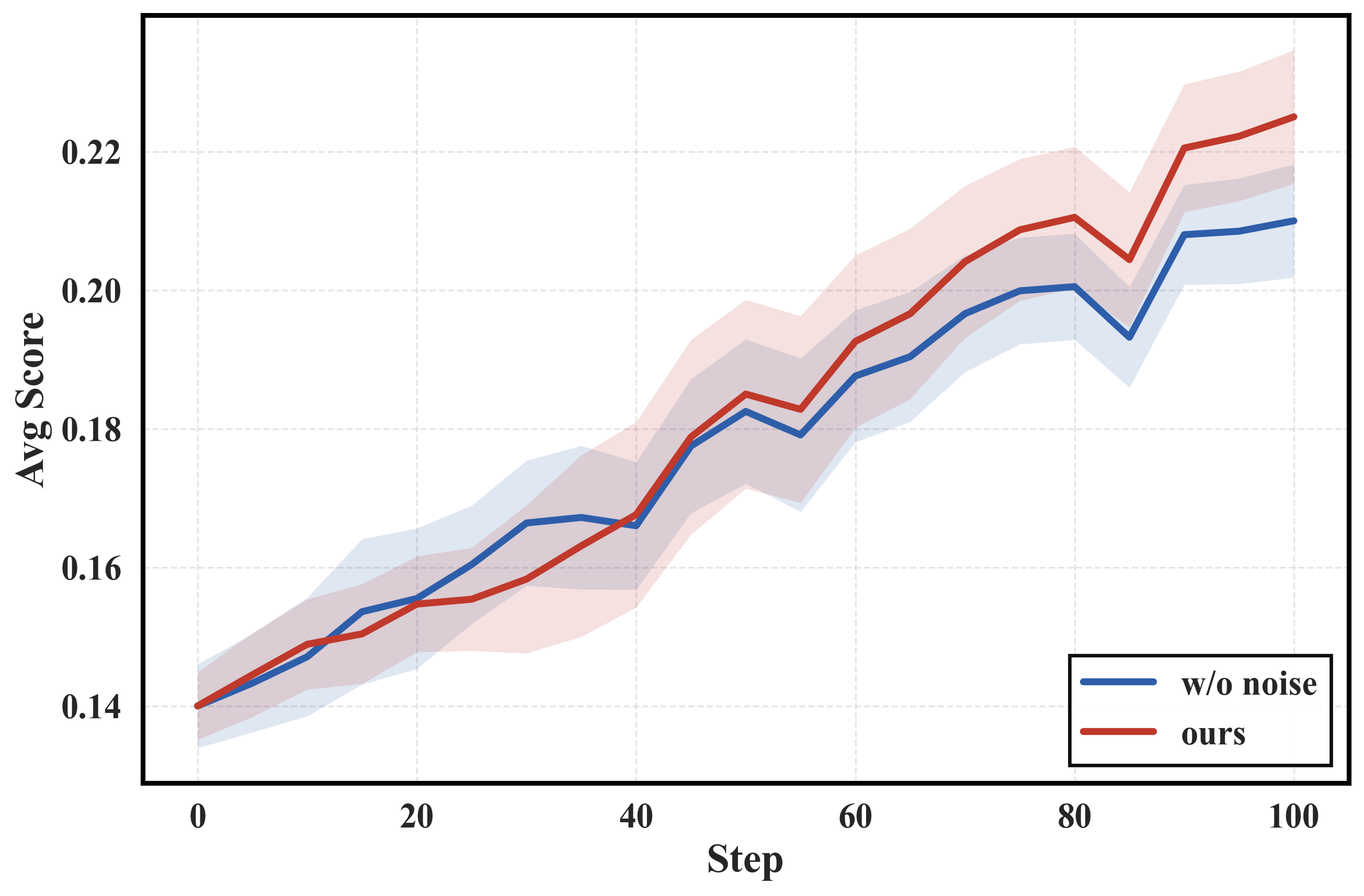}
    \caption{Idealized setting}
    \label{fig:train_no_noise}
\end{subfigure}
\hfill
\begin{subfigure}[t]{0.45\columnwidth}
    \centering
    \includegraphics[width=\linewidth]{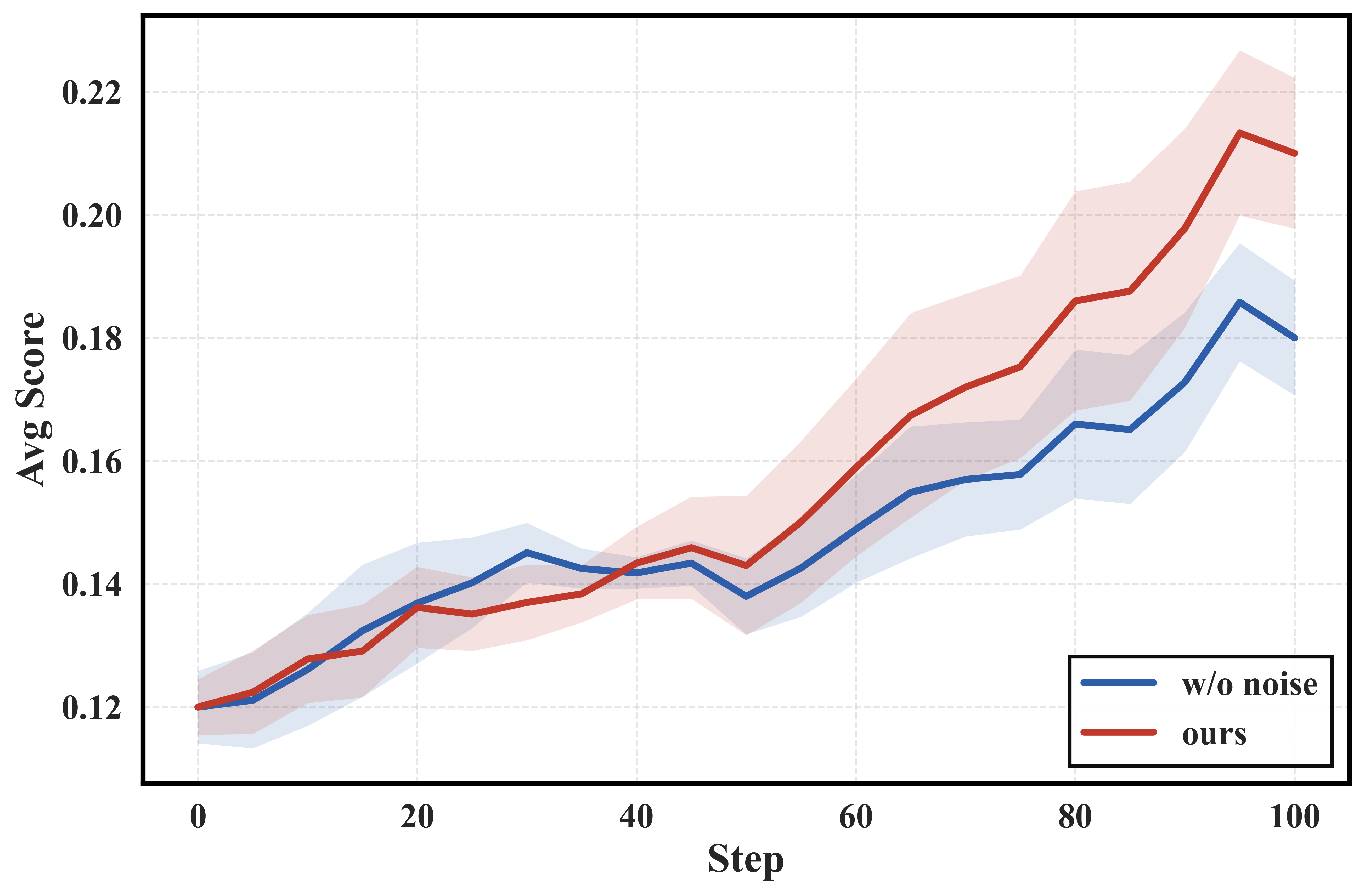}
    \caption{Noisy setting}
    \label{fig:train_noise}
\end{subfigure}
\caption{Training dynamics on Vita-Bench Delivery (Qwen3-8B). We compare \ours with a baseline trained without noise under both ideal (no-noise) and noisy evaluations.}
\label{fig:training_dynamics}
\end{figure}

Figure~\ref{fig:training_dynamics} compares the training dynamics of \ours and the baseline trained without noise under both ideal and noisy evaluations. 
In the early stage of training, the two methods exhibit comparable performance, as optimization is largely conducted on clean trajectories serving as a warm-up phase. 
The initial introduction of moderate noise may even lead to a slight degradation in performance, reflecting the increased difficulty of the perturbed trajectories. 
As training progresses, the model gradually adapts to noisy conditions, and the curriculum introduces increasingly challenging perturbations, raising the requirements for successful task completion. 
While the baseline continues to improve, its gains remain moderate. In contrast, \ours achieves more substantial improvements, particularly under the noisy evaluation, where the performance gap becomes increasingly pronounced. 
This trend indicates that learning in noisy environments provides informative training signals, enabling the agent to develop stronger robustness and improved performance under challenging conditions.

\paragraph{Interaction Pattern.}

Beyond aggregate performance, we analyze how curriculum training alters the agent's interaction behavior compared to the base model and GSPO, along three dimensions: tool usage, response verbosity, and reasoning overhead.
As shown in Table~\ref{tab:interaction_pattern}, under the noisy setting, \ours reduces tool usage from 13.9 to 11.4 calls per episode (18\%), while GSPO yields only marginal change. In contrast, under the ideal setting, all methods exhibit similar tool usage (6.7--7.4 calls), with negligible differences. This indicates that the reduction in tool calls is not due to a general degradation of capability, but arises specifically in noisy environments, where \ours avoids excessive or redundant interactions.
In parallel, \ours produces substantially longer responses, with output tokens increasing from 2{,}014 to 4{,}248 under noise (2.1$\times$), and a consistent trend observed in the ideal setting. This suggests a shift toward more explicit and detailed interaction, potentially reducing the need for additional clarification through further tool calls.
Taken together, these results show that curriculum training primarily improves the efficiency and clarity of interaction—reducing unnecessary tool usage while producing more informative responses.
We provide a case study in Appendix~\ref{app:case_study}.

\section{Related Work} 
\label{sec:related-work}

\subsection{LLM as Agent}
\label{sec:related-llm-agent}

With the rapid improvement in reasoning and instruction-following capabilities, LLMs have evolved from passive text generators into agents capable of tool use, multi-step planning, and interaction with dynamic environments~\citep{yao2023react,schick2023toolformer,shinn2023reflexion,wang2023voyager}. 
Early approaches primarily rely on hand-crafted pipelines, where reasoning–action patterns, tool schemas, and memory mechanisms are manually designed on top of frozen models \citep{yao2023react,shinn2023reflexion,wu2023autogen,hong2024metagpt,park2023generative}. 
While effective, such prompt-level designs are brittle and do not fundamentally improve the underlying policy.
More recent work instead trains agent behaviors directly via reinforcement learning with verifiable rewards (RLVR)~\citep{guo2025deepseekr1,lambert2024tulu3,shao2024deepseekmath}. 
One line of research focuses on stabilizing long-horizon training and improving credit assignment, with a variety of algorithmic advances~\citep{shao2024deepseekmath,yu2025dapo,
zheng2025gspo,liu2025drgrpo,yuan2025vapo,yao2026coba}. 
In parallel, another line of work explores scalable environment design and task construction, enabling RL training over increasingly diverse and realistic agent scenarios, including tool use and retrieval~\citep{feng2025retool,jin2025searchr1,song2025r1searcher,shi2025look}, software engineering~\citep{jimenez2024swebench,pan2024swegym,wei2025swerl}, and web or GUI interaction~\citep{zhou2024webarena,xie2024osworld,trivedi2024appworld}.
Despite these advances, existing work is largely conducted under idealized settings, leaving a gap between training conditions and real-world noisy environments.

\begin{table}[t]
\centering
\small
\setlength{\tabcolsep}{4pt}
\caption{Interaction patterns on Retail domain with Qwen3-8B.}
\label{tab:interaction_pattern}
\begin{tabular}{lccc}
\toprule
Method & Tool Calls & Output Tokens & Reasoning Tokens \\
\midrule
\multicolumn{4}{l}{\textit{Noisy setting:}} \\
Base        & 13.9 & 2{,}014 & 10{,}897 \\
GSPO        & 13.7 & 2{,}180 & 11{,}012 \\
Ours        & 11.4 & 4{,}248 & 10{,}964 \\
\midrule
\multicolumn{4}{l}{\textit{Ideal setting:}} \\
Base        & 7.1 & 1{,}931 & 7{,}091 \\
GSPO        & 7.4 & 1{,}982 & 7{,}265 \\
Ours        & 6.7 & 3{,}923 & 7{,}534 \\
\bottomrule
\end{tabular}
\end{table}

\subsection{Robustness of Agent}
\label{sec:related-robustness}

As LLM-based agents are increasingly deployed in complex real-world settings, \emph{robustness} has emerged as a critical concern alongside raw capability~\citep{li2025robustcode,levy2025advrobust,agrawal2025perturbinstr}. 
A growing body of work shows that agent performance degrades substantially under distributional shifts in environment dynamics~\citep{anghel2025diagbias,wan2025robustllm,herrerapoyatos2025uncertainty,wang2025materials,xue2025illusion,yu2025reasonrobust}. 
On the user side, prior work investigates how perturbations in prompts, clarifications, and multi-turn dialog interactions affect agent behavior~\citep{li2025structflow,deshpande2025multichallenge,qi2025agentif,wang2024understandingux,gan2024clarqllm,zhang2024clamber,yang2025whatpromptsdontsay}. 
These studies suggest that realistic user interactions are often noisy, under-specified, and evolving, exposing agents to a broader and more dynamic input distribution than curated settings. 
On the \emph{execution side}, reliance on external tools introduces an additional source of instability, as tools may return incomplete, outdated, or erroneous outputs~\citep{xu2024reliability,zhang2024toolbehonest,vuddanti2025paladin,kokane2025toolscan,xiong2025butterfly,zhan2024injecagent,zhang2025adversarial}. 
Such local errors frequently propagate along the interaction trajectory, leading to cascading failures in downstream decisions~\citep{yang2025whatpromptsdontsay,zhu2025tooluse,song2024trialerror}. 
To systematically characterize these effects, recent work proposes robustness benchmarks and diagnostic protocols~\citep{nalbandyan2025score,wen2025scenarioindep,yu2025reasonrobust,lunardi2025reliability,siska2024inadequacy}. 
AgentNoiseBench~\citep{wang2026agentnoisebench} further introduces a unified taxonomy of user-side and tool-side noise with controllable perturbations, revealing consistent performance degradation across a wide range of models under realistic noise. 
However, existing approaches primarily focus on evaluation, leaving the problem of learning robust agent behaviors under realistic noise largely underexplored.
\section{Limitation} 
\label{sec:limitation}

While our framework demonstrates consistent improvements in robustness, we note several aspects that could be further explored in future work.
First, our primary goal is to investigate whether incorporating real-world interaction noise can improve the robustness of agent policies. 
To this end, we focus on two representative sources of noise—user-side and tool-side perturbations—and model a set of common failure patterns observed in practice. 
While this design captures a broad range of realistic imperfections, it does not aim to exhaustively cover all possible forms of uncertainty. 
In real-world environments, noise can be more complex, compositional, and dynamically evolving. 
Extending the framework to model richer and more diverse interaction patterns is an important direction for future work.
Also, our experiments are primarily conducted in synthesized environments that approximate real-world interaction dynamics. 
In principle, the proposed framework is general and can be applied to any agentic environment by augmenting it with structured noise. 
However, due to the high cost of agentic training and the need to systematically evaluate robustness under out-of-distribution conditions, we focus on controlled settings rather than extensively benchmarking across multiple in-domain training and testing datasets. 
We believe that applying our framework to broader real-world and in-domain benchmarks is an important direction for future work.
We leave these directions for future work.
\section{Conclusion}
\label{sec:conclusion}

In this work, we investigate the fundamental gap between idealized agentic training and real-world deployment, and identify the lack of environmental imperfections during training as a key factor limiting agent robustness. 
To address this issue, we propose a noise-aware training framework that explicitly incorporates stochasticity and imperfections into the agent learning process. 
By systematically modeling instruction noise and tool noise, and introducing them through an automatic noise injection pipeline, our approach exposes agents to more realistic interaction dynamics. 
To ensure stable optimization, we further design an adaptive training strategy that combines clean and perturbed rollouts while progressively increasing noise difficulty based on the model’s robustness.
Extensive experiments demonstrate that our method consistently improves agent performance under noisy and dynamic environments, validating its effectiveness in enhancing robustness. 
Notably, we also observe consistent gains on standard, idealized benchmarks, suggesting that controlled exposure to environmental noise promotes more generalizable reasoning and decision-making behaviors. 
Overall, this work highlights the importance of aligning training conditions with real-world interaction characteristics, and provides a practical framework for improving the robustness of LLM-based agents in realistic deployment settings.
% \begin{ack}

% \end{ack}

\bibliographystyle{unsrtnat}
\bibliography{neurips_2024}

@article{nalbandyan2025score,
  title={SCORE: Systematic COnsistency and Robustness Evaluation for Large Language Models},
  author={Nalbandyan, Grigor and Shahbazyan, Rima and Bakhturina, Evelina},
  journal={arXiv preprint arXiv:2503.00137},
  year={2025}
}

@inproceedings{kokane2025toolscan,
  title={ToolScan: A Benchmark For Characterizing Errors In Tool-Use LLMs},
  author={Kokane, Shirley and Zhu, Ming and Awalgaonkar, Tulika Manoj and Zhang, Jianguo and Prabhakar, Akshara and Hoang, Thai Quoc and Liu, Zuxin and RN, Rithesh and Yang, Liangwei and Yao, Weiran and others},
  booktitle={ICLR 2025 Workshop on Building Trust in Language Models and Applications}
}

@article{zhang2024toolbehonest,
  title={Toolbehonest: A multi-level hallucination diagnostic benchmark for tool-augmented large language models},
  author={Zhang, Yuxiang and Chen, Jing and Wang, Junjie and Liu, Yaxin and Yang, Cheng and Shi, Chufan and Zhu, Xinyu and Lin, Zihao and Wan, Hanwen and Yang, Yujiu and others},
  journal={arXiv preprint arXiv:2406.20015},
  year={2024}
}

@article{zhan2024injecagent,
  title={Injecagent: Benchmarking indirect prompt injections in tool-integrated large language model agents},
  author={Zhan, Qiusi and Liang, Zhixiang and Ying, Zifan and Kang, Daniel},
  journal={arXiv preprint arXiv:2403.02691},
  year={2024}
}

@article{zhang2024clamber,
  title={Clamber: A benchmark of identifying and clarifying ambiguous information needs in large language models},
  author={Zhang, Tong and Qin, Peixin and Deng, Yang and Huang, Chen and Lei, Wenqiang and Liu, Junhong and Jin, Dingnan and Liang, Hongru and Chua, Tat-Seng},
  journal={arXiv preprint arXiv:2405.12063},
  year={2024}
}

@article{gemini3pro,
  title={Gemini 3 Pro Model Card},
  author={Google},
  journal={https://storage.googleapis.com/deepmind-media/Model-Cards/Gemini-3-Pro-Model-Card.pdf},
  year={2025}
}

@article{team2025longcat-flash-thinking,
  title={Introducing LongCat-Flash-Thinking: A Technical Report},
  author={Team, Meituan LongCat and Gui, Anchun and Li, Bei and Tao, Bingyang and Zhou, Bole and Chen, Borun and Zhang, Chao and Han, Chengcheng and Yang, Chenhui and Zhang, Chi and others},
  journal={arXiv preprint arXiv:2509.18883},
  year={2025}
}

@article{qwen3-32B,
  title={https://qwen.ai/blog?id=qwen3},
  author={Qwen Team},
  year         = {2025},
  url          = {https://qwen.ai/blog?id=qwen3-max}
}

@article{liu2025deepseekv32,
  title={Deepseek-v3. 2: Pushing the frontier of open large language models},
  author={Liu, Aixin and Mei, Aoxue and Lin, Bangcai and Xue, Bing and Wang, Bingxuan and Xu, Bingzheng and Wu, Bochao and Zhang, Bowei and Lin, Chaofan and Dong, Chen and others},
  journal={arXiv preprint arXiv:2512.02556},
  year={2025}
}

@article{guo2025deepseekr1,
  title={Deepseek-r1: Incentivizing reasoning capability in llms via reinforcement learning},
  author={Guo, Daya and Yang, Dejian and Zhang, Haowei and Song, Junxiao and Zhang, Ruoyu and Xu, Runxin and Zhu, Qihao and Ma, Shirong and Wang, Peiyi and Bi, Xiao and others},
  journal={arXiv preprint arXiv:2501.12948},
  year={2025}
}

@article{gpt52,
  author       = {{OpenAI}},
  title        = {Introducing GPT-5.2},
  year         = {2025},
  url          = {https://openai.com/index/introducing-gpt-5-2/},
}

@article{xiong2025butterfly,
  title={Butterfly effects in toolchains: A comprehensive analysis of failed parameter filling in llm tool-agent systems},
  author={Xiong, Qian and Huang, Yuekai and Jiang, Ziyou and Chang, Zhiyuan and Zheng, Yujia and Li, Tianhao and Li, Mingyang},
  journal={arXiv preprint arXiv:2507.15296},
  year={2025}
}

@article{rawles2024androidworld,
  title={Androidworld: A dynamic benchmarking environment for autonomous agents},
  author={Rawles, Christopher and Clinckemaillie, Sarah and Chang, Yifan and Waltz, Jonathan and Lau, Gabrielle and Fair, Marybeth and Li, Alice and Bishop, William and Li, Wei and Campbell-Ajala, Folawiyo and others},
  journal={arXiv preprint arXiv:2405.14573},
  year={2024}
}

@inproceedings{sun2025gui,
  title={Gui-xplore: Empowering generalizable gui agents with one exploration},
  author={Sun, Yuchen and Zhao, Shanhui and Yu, Tao and Wen, Hao and Va, Samith and Xu, Mengwei and Li, Yuanchun and Zhang, Chongyang},
  booktitle={Proceedings of the Computer Vision and Pattern Recognition Conference},
  pages={19477--19486},
  year={2025}
}

@inproceedings{shoeb2025out,
  title={Out-of-distribution segmentation in autonomous driving: Problems and state of the art},
  author={Shoeb, Youssef and Nowzad, Azarm and Gottschalk, Hanno},
  booktitle={Proceedings of the Computer Vision and Pattern Recognition Conference},
  pages={4310--4320},
  year={2025}
}

@article{gallois2005communication,
  title={Communication accommodation theory},
  author={Gallois, Cindy and Ogay, Tania and Giles, Howard},
  journal={Theorizing about intercultural communication},
  pages={121--148},
  year={2005}
}

@inproceedings{trippas2024users,
  title={What do users really ask large language models? an initial log analysis of google bard interactions in the wild},
  author={Trippas, Johanne R and Al Lawati, Sara Fahad Dawood and Mackenzie, Joel and Gallagher, Luke},
  booktitle={Proceedings of the 47th International ACM SIGIR Conference on Research and Development in Information Retrieval},
  pages={2703--2707},
  year={2024}
}

@article{wang2024understanding,
  title={Understanding user experience in large language model interactions},
  author={Wang, Jiayin and Ma, Weizhi and Sun, Peijie and Zhang, Min and Nie, Jian-Yun},
  journal={arXiv preprint arXiv:2401.08329},
  year={2024}
}

@inproceedings{zeng2024agenttuning,
  title={Agenttuning: Enabling generalized agent abilities for llms},
  author={Zeng, Aohan and Liu, Mingdao and Lu, Rui and Wang, Bowen and Liu, Xiao and Dong, Yuxiao and Tang, Jie},
  booktitle={Findings of the Association for Computational Linguistics: ACL 2024},
  pages={3053--3077},
  year={2024}
}

@article{qi2024webrl,
  title={Webrl: Training llm web agents via self-evolving online curriculum reinforcement learning},
  author={Qi, Zehan and Liu, Xiao and Iong, Iat Long and Lai, Hanyu and Sun, Xueqiao and Zhao, Wenyi and Yang, Yu and Yang, Xinyue and Sun, Jiadai and Yao, Shuntian and others},
  journal={arXiv preprint arXiv:2411.02337},
  year={2024}
}

@article{jin2025search,
  title={Search-r1: Training llms to reason and leverage search engines with reinforcement learning},
  author={Jin, Bowen and Zeng, Hansi and Yue, Zhenrui and Yoon, Jinsung and Arik, Sercan and Wang, Dong and Zamani, Hamed and Han, Jiawei},
  journal={arXiv preprint arXiv:2503.09516},
  year={2025}
}

@article{xue2025illusion,
  title={An illusion of progress? assessing the current state of web agents},
  author={Xue, Tianci and Qi, Weijian and Shi, Tianneng and Song, Chan Hee and Gou, Boyu and Song, Dawn and Sun, Huan and Su, Yu},
  journal={arXiv preprint arXiv:2504.01382},
  year={2025}
}

@article{deng2023mind2web,
  title={Mind2web: Towards a generalist agent for the web},
  author={Deng, Xiang and Gu, Yu and Zheng, Boyuan and Chen, Shijie and Stevens, Sam and Wang, Boshi and Sun, Huan and Su, Yu},
  journal={Advances in Neural Information Processing Systems},
  volume={36},
  pages={28091--28114},
  year={2023}
}

@article{zhou2023webarena,
  title={Webarena: A realistic web environment for building autonomous agents},
  author={Zhou, Shuyan and Xu, Frank F and Zhu, Hao and Zhou, Xuhui and Lo, Robert and Sridhar, Abishek and Cheng, Xianyi and Ou, Tianyue and Bisk, Yonatan and Fried, Daniel and others},
  journal={arXiv preprint arXiv:2307.13854},
  year={2023}
}

@article{team2025kimi,
  title={Kimi k2: Open agentic intelligence},
  author={Team, Kimi and Bai, Yifan and Bao, Yiping and Chen, Guanduo and Chen, Jiahao and Chen, Ningxin and Chen, Ruijue and Chen, Yanru and Chen, Yuankun and Chen, Yutian and others},
  journal={arXiv preprint arXiv:2507.20534},
  year={2025}
}

@article{zeng2025glm,
  title={Glm-4.5: Agentic, reasoning, and coding (arc) foundation models},
  author={Zeng, Aohan and Lv, Xin and Zheng, Qinkai and Hou, Zhenyu and Chen, Bin and Xie, Chengxing and Wang, Cunxiang and Yin, Da and Zeng, Hao and Zhang, Jiajie and others},
  journal={arXiv preprint arXiv:2508.06471},
  year={2025}
}

@article{team2025longcat,
  title={Longcat-flash technical report},
  author={Team, Meituan LongCat and Li, Bei and Lei, Bingye and Wang, Bo and Rong, Bolin and Wang, Chao and Zhang, Chao and Gao, Chen and Zhang, Chen and Sun, Cheng and others},
  journal={arXiv preprint arXiv:2509.01322},
  year={2025}
}

@article{vuddanti2025paladin,
  title={PALADIN: Self-Correcting Language Model Agents to Cure Tool-Failure Cases},
  author={Vuddanti, Sri Vatsa and Shah, Aarav and Chittiprolu, Satwik Kumar and Song, Tony and Dev, Sunishchal and Zhu, Kevin and Chaudhary, Maheep},
  journal={arXiv preprint arXiv:2509.25238},
  year={2025}
}

@article{qi2025agentif,
  title={Agentif: Benchmarking instruction following of large language models in agentic scenarios},
  author={Qi, Yunjia and Peng, Hao and Wang, Xiaozhi and Xin, Amy and Liu, Youfeng and Xu, Bin and Hou, Lei and Li, Juanzi},
  journal={arXiv preprint arXiv:2505.16944},
  year={2025}
}

@inproceedings{deshpande2025multichallenge,
  title={Multichallenge: A realistic multi-turn conversation evaluation benchmark challenging to frontier llms},
  author={Deshpande, Kaustubh and Sirdeshmukh, Ved and Mols, Johannes Baptist and Jin, Lifeng and Hernandez-Cardona, Ed-Yeremai and Lee, Dean and Kritz, Jeremy and Primack, Willow E and Yue, Summer and Xing, Chen},
  booktitle={Findings of the Association for Computational Linguistics: ACL 2025},
  pages={18632--18702},
  year={2025}
}

@article{schick2023toolformer,
  title={Toolformer: Language models can teach themselves to use tools},
  author={Schick, Timo and Dwivedi-Yu, Jane and Dess{\`\i}, Roberto and Raileanu, Roberta and Lomeli, Maria and Hambro, Eric and Zettlemoyer, Luke and Cancedda, Nicola and Scialom, Thomas},
  journal={Advances in Neural Information Processing Systems},
  volume={36},
  pages={68539--68551},
  year={2023}
}

@article{he2025vitabench,
  title={Vitabench: Benchmarking llm agents with versatile interactive tasks in real-world applications},
  author={He, Wei and Sun, Yueqing and Hao, Hongyan and Hao, Xueyuan and Xia, Zhikang and Gu, Qi and Han, Chengcheng and Zhao, Dengchang and Su, Hui and Zhang, Kefeng and others},
  journal={arXiv preprint arXiv:2509.26490},
  year={2025}
}

@article{yao2024tau,
  author       = {Shunyu Yao and
                  Noah Shinn and
                  Pedram Razavi and
                  Karthik Narasimhan},
  title        = {{\(\tau\)}-bench: {A} Benchmark for Tool-Agent-User Interaction in
                  Real-World Domains},
  journal      = {CoRR},
  volume       = {abs/2406.12045},
  year         = {2024},
  url          = {https://doi.org/10.48550/arXiv.2406.12045},
  doi          = {10.48550/ARXIV.2406.12045},
  eprinttype    = {arXiv},
  eprint       = {2406.12045},
  bibsource    = {dblp computer science bibliography, https://dblp.org}
}

@article{barres2025tau,
  title={$\tau^2$-Bench: Evaluating Conversational Agents in a Dual-Control Environment},
  author={Barres, Victor and Dong, Honghua and Ray, Soham and Si, Xujie and Narasimhan, Karthik},
  journal={arXiv preprint arXiv:2506.07982},
  year={2025}
}

@inproceedings{tobin2017domain,
  title     = {Domain Randomization for Transferring Deep Neural Networks from Simulation to the Real World},
  author    = {Tobin, Josh and Fong, Rachel and Ray, Alex and Schneider, Jonas and Zaremba, Wojciech and Abbeel, Pieter},
  booktitle = {IEEE/RSJ International Conference on Intelligent Robots and Systems (IROS)},
  year      = {2017}
}

@inproceedings{sadeghi2017cad2rl,
  title     = {CAD2RL: Real Single-Image Flight Without a Single Real Image},
  author    = {Sadeghi, Fereshteh and Levine, Sergey},
  booktitle = {Robotics: Science and Systems (RSS)},
  year      = {2017}
}

@inproceedings{zhao2021robust,
  title     = {Robust Reinforcement Learning as a Stackelberg Game},
  author    = {Zhao, Minghao and Xiong, Wenhan and Zhang, Lei and others},
  booktitle = {International Conference on Machine Learning (ICML)},
  year      = {2021}
}

@article{agentrl_survey2025,
  title={The Landscape of Agentic Reinforcement Learning for {LLM}s: A Survey},
  author={Zhao, Hao and others},
  journal={arXiv preprint arXiv:2509.02547},
  year={2025}
}

@article{deepseek_r1,
  title={DeepSeek-{R1}: Incentivizing Reasoning Capability in {LLM}s via Reinforcement Learning},
  author={DeepSeek-AI and Guo, Daya and Yang, Dejian and Zhang, Haowei and Song, Junxiao and Zhang, Ruoyu and Xu, Runxin and Zhu, Qihao and Ma, Shirong and Wang, Peiyi and others},
  journal={arXiv preprint arXiv:2501.12948},
  year={2025}
}

@article{verltool2025,
  title={VerlTool: Towards Holistic Agentic Reinforcement Learning with Tool Use},
  author={Jiang, Qiushi and others},
  journal={arXiv preprint arXiv:2509.01055},
  year={2025}
}

@article{shao2024grpo,
  title={DeepSeekMath: Pushing the Limits of Mathematical Reasoning in Open Language Models},
  author={Shao, Zhihong and Wang, Peiyi and Zhu, Qihao and Xu, Runxin and Song, Junxiao and Bi, Xiao and Zhang, Haowei and Zhang, Mingchuan and Li, Y. K. and Wu, Yang and others},
  journal={arXiv preprint arXiv:2402.03300},
  year={2024}
}

@article{scaleenv2025,
  title={ScaleEnv: Scaling Environment Synthesis from Scratch for Generalist Interactive Tool-Use Agent Training},
  author={Tu, Dunwei and Hao, Hongyan and Yang, Hansi and Chen, Yihao and Zhang, Yi-Kai and Xia, Zhikang and Yang, Yu and Sun, Yueqing and Liu, Xingchen and Shen, Furao and Gu, Qi and Su, Hui and Cai, Xunliang},
  journal={arXiv preprint arXiv:2602.06820},
  year={2026}
}

@article{schulman2017ppo,
  title   = {Proximal Policy Optimization Algorithms},
  author  = {Schulman, John and Wolski, Filip and Dhariwal, Prafulla and Radford, Alec and Klimov, Oleg},
  journal = {arXiv preprint arXiv:1707.06347},
  year    = {2017}
}

@inproceedings{yao2023react,
  title={ReAct: Synergizing Reasoning and Acting in Language Models},
  author={Yao, Shunyu and Zhao, Jeffrey and Yu, Dian and Du, Nan and Shafran, Izhak and Narasimhan, Karthik and Cao, Yuan},
  booktitle={International Conference on Learning Representations (ICLR)},
  year={2023}
}

@inproceedings{shinn2023reflexion,
  title={Reflexion: Language Agents with Verbal Reinforcement Learning},
  author={Shinn, Noah and Cassano, Federico and Berman, Edward and Gopinath, Ashwin and Narasimhan, Karthik and Yao, Shunyu},
  booktitle={Advances in Neural Information Processing Systems (NeurIPS)},
  year={2023}
}

@article{wang2023voyager,
  title={Voyager: An Open-Ended Embodied Agent with Large Language Models},
  author={Wang, Guanzhi and Xie, Yuqi and Jiang, Yunfan and Mandlekar, Ajay and Xiao, Chaowei and Zhu, Yuke and Fan, Linxi and Anandkumar, Anima},
  journal={arXiv preprint arXiv:2305.16291},
  year={2023}
}

@article{wu2023autogen,
  title={AutoGen: Enabling Next-Gen LLM Applications via Multi-Agent Conversation},
  author={Wu, Qingyun and Bansal, Gagan and Zhang, Jieyu and Wu, Yiran and Li, Beibin and Zhu, Erkang and Jiang, Li and Zhang, Xiaoyun and Zhang, Shaokun and Liu, Jiale and others},
  journal={arXiv preprint arXiv:2308.08155},
  year={2023}
}

@inproceedings{hong2024metagpt,
  title={MetaGPT: Meta Programming for A Multi-Agent Collaborative Framework},
  author={Hong, Sirui and Zhuge, Mingchen and Chen, Jonathan and Zheng, Xiawu and Cheng, Yuheng and Zhang, Ceyao and Wang, Jinlin and Wang, Zili and Yau, Steven Ka Shing and Lin, Zijuan and others},
  booktitle={International Conference on Learning Representations (ICLR)},
  year={2024}
}

@inproceedings{park2023generative,
  title={Generative Agents: Interactive Simulacra of Human Behavior},
  author={Park, Joon Sung and O'Brien, Joseph C and Cai, Carrie J and Morris, Meredith Ringel and Liang, Percy and Bernstein, Michael S},
  booktitle={ACM Symposium on User Interface Software and Technology (UIST)},
  year={2023}
}

@article{lambert2024tulu3,
  title={T{\"u}lu 3: Pushing Frontiers in Open Language Model Post-Training},
  author={Lambert, Nathan and Morrison, Jacob and Pyatkin, Valentina and Huang, Shengyi and Ivison, Hamish and Brahman, Faeze and Miranda, Lester James V. and Liu, Alisa and Dziri, Nouha and Lyu, Shane and others},
  journal={arXiv preprint arXiv:2411.15124},
  year={2024}
}

@article{shao2024deepseekmath,
  title={DeepSeekMath: Pushing the Limits of Mathematical Reasoning in Open Language Models},
  author={Shao, Zhihong and Wang, Peiyi and Zhu, Qihao and Xu, Runxin and Song, Junxiao and Bi, Xiao and Zhang, Haowei and Zhang, Mingchuan and Li, Y.K. and Wu, Y. and Guo, Daya},
  journal={arXiv preprint arXiv:2402.03300},
  year={2024}
}

@article{yu2025dapo,
  title={{DAPO}: An Open-Source {LLM} Reinforcement Learning System at Scale},
  author={Yu, Qiying and Zhang, Zheng and Zhu, Ruofei and Yuan, Yufeng and Zuo, Xiaochen and Yue, Yu and Fan, Tiantian and Liu, Gaohong and Liu, Lingjun and Liu, Xin and others},
  journal={arXiv preprint arXiv:2503.14476},
  year={2025}
}

@article{zheng2025gspo,
  title={Group Sequence Policy Optimization},
  author={Zheng, Chujie and Liu, Shixuan and Li, Mingze and Chen, Xiong-Hui and Yu, Bowen and Gao, Chang and Dang, Kai and Liu, Yuqiong and Men, Rui and Yang, An and Zhou, Jingren and Lin, Junyang},
  journal={arXiv preprint arXiv:2507.18071},
  year={2025}
}

@article{liu2025drgrpo,
  title={Understanding {R1}-Zero-Like Training: A Critical Perspective},
  author={Liu, Zichen and Chen, Changyu and Li, Wenjun and Qi, Penghui and Pang, Tianyu and Du, Chao and Lee, Wee Sun and Lin, Min},
  journal={arXiv preprint arXiv:2503.20783},
  year={2025}
}

@article{yuan2025vapo,
  title={{VAPO}: Efficient and Reliable Reinforcement Learning for Advanced Reasoning Tasks},
  author={Yuan, Yufeng and Yu, Qiying and Zuo, Xiaochen and Zhu, Ruofei and Xu, Wenyuan and Chen, Jiaze and Wang, Chengyi and Fan, Tiantian and Du, Zhengyin and Yan, Xiangpeng and others},
  journal={arXiv preprint arXiv:2504.05118},
  year={2025}
}

@article{feng2025retool,
  title={{ReTool}: Reinforcement Learning for Strategic Tool Use in {LLMs}},
  author={Feng, Jiazhan and Huang, Shijue and Qu, Xingwei and Zhang, Ge and Qin, Yujia and Zhong, Baoquan and Jiang, Chengquan and Chi, Jinxin and Zhong, Wanjun},
  journal={arXiv preprint arXiv:2504.11536},
  year={2025}
}

@article{jin2025searchr1,
  title={{Search-R1}: Training {LLMs} to Reason and Leverage Search Engines with Reinforcement Learning},
  author={Jin, Bowen and Zeng, Hansi and Yue, Zhenrui and Yoon, Jinsung and Arik, Sercan and Wang, Dong and Zamani, Hamed and Han, Jiawei},
  journal={arXiv preprint arXiv:2503.09516},
  year={2025}
}

@article{song2025r1searcher,
  title={{R1-Searcher}: Incentivizing the Search Capability in {LLMs} via Reinforcement Learning},
  author={Song, Huatong and Jiang, Jinhao and Min, Yingqian and Chen, Jie and Chen, Zhipeng and Zhao, Wayne Xin and Fang, Lei and Wen, Ji-Rong},
  journal={arXiv preprint arXiv:2503.05592},
  year={2025}
}

@inproceedings{jimenez2024swebench,
  title={{SWE-bench}: Can Language Models Resolve Real-World {GitHub} Issues?},
  author={Jimenez, Carlos E. and Yang, John and Wettig, Alexander and Yao, Shunyu and Pei, Kexin and Press, Ofir and Narasimhan, Karthik},
  booktitle={International Conference on Learning Representations (ICLR)},
  year={2024}
}

@article{pan2024swegym,
  title={Training Software Engineering Agents and Verifiers with {SWE-Gym}},
  author={Pan, Jiayi and Wang, Xingyao and Neubig, Graham and Jaitly, Navdeep and Ji, Heng and Suhr, Alane and Zhang, Yizhe},
  journal={arXiv preprint arXiv:2412.21139},
  year={2024}
}

@article{wei2025swerl,
  title={{SWE-RL}: Advancing {LLM} Reasoning via Reinforcement Learning on Open Software Evolution},
  author={Wei, Yuxiang and Duchenne, Olivier and Copet, Jade and Carbonneaux, Quentin and Zhang, Lingming and Fried, Daniel and Synnaeve, Gabriel and Singh, Rishabh and Wang, Sida I.},
  journal={arXiv preprint arXiv:2502.18449},
  year={2025}
}

@inproceedings{zhou2024webarena,
  title={{WebArena}: A Realistic Web Environment for Building Autonomous Agents},
  author={Zhou, Shuyan and Xu, Frank F. and Zhu, Hao and Zhou, Xuhui and Lo, Robert and Sridhar, Abishek and Cheng, Xianyi and Ou, Tianyue and Bisk, Yonatan and Fried, Daniel and Alon, Uri and Neubig, Graham},
  booktitle={International Conference on Learning Representations (ICLR)},
  year={2024}
}

@inproceedings{xie2024osworld,
  title={{OSWorld}: Benchmarking Multimodal Agents for Open-Ended Tasks in Real Computer Environments},
  author={Xie, Tianbao and Zhang, Danyang and Chen, Jixuan and Li, Xiaochuan and Zhao, Siheng and Cao, Ruisheng and Hua, Toh Jing and Cheng, Zhoujun and Shin, Dongchan and Lei, Fangyu and others},
  booktitle={Advances in Neural Information Processing Systems (NeurIPS)},
  year={2024}
}

@inproceedings{trivedi2024appworld,
  title={{AppWorld}: A Controllable World of Apps and People for Benchmarking Interactive Coding Agents},
  author={Trivedi, Harsh and Khot, Tushar and Hartmann, Mareike and Manku, Ruskin and Dong, Vinty and Li, Edward and Gupta, Shashank and Sabharwal, Ashish and Balasubramanian, Niranjan},
  booktitle={Annual Meeting of the Association for Computational Linguistics (ACL)},
  year={2024}
}

@article{li2025robustcode,
  title={Enhancing the Robustness of {LLM}-Generated Code: Empirical Study and Framework},
  author={Li, Zike and Liu, Mingwei and Li, An and He, Kaifeng and Wang, Yanlin and Peng, Xin and Zheng, Zibin},
  journal={arXiv preprint arXiv:2503.20197},
  year={2025}
}

@article{levy2025advrobust,
  title={Towards Robust {LLMs}: An Adversarial Robustness Measurement Framework},
  author={Levy, Natan and Ashrov, Adiel and Katz, Guy},
  journal={arXiv preprint arXiv:2504.17723},
  year={2025}
}

@article{agrawal2025perturbinstr,
  title={Enhancing {LLM} Robustness to Perturbed Instructions: An Empirical Study},
  author={Agrawal, Aryan and Alazraki, Lisa and Honarvar, Shahin and Rei, Marek},
  journal={arXiv preprint arXiv:2504.02733},
  year={2025}
}

@article{anghel2025diagbias,
  title={Diagnosing Bias and Instability in {LLM} Evaluation: A Scalable Pairwise Meta-Evaluator},
  author={Anghel, Catalin and Anghel, Andreea A. and Pecheanu, Emilia and Cocu, Adina and Istrate, Adrian and Andrei, Constantin A.},
  journal={Information},
  volume={16},
  number={8},
  pages={652},
  year={2025}
}

@article{wan2025robustllm,
  title={Robust {LLM} Training Infrastructure at {ByteDance}},
  author={Wan, Borui and Liu, Gaohong and Song, Zhe and Wang, Jiarui and Zhang, Yukang and Sheng, Guangming and Wang, Shuguang and Wei, Hui and Wang, Chao and Lou, Wen and others},
  booktitle={ACM SIGOPS 31st Symposium on Operating Systems Principles (SOSP)},
  pages={186--203},
  year={2025}
}

@article{herrerapoyatos2025uncertainty,
  title={An Overview of Model Uncertainty and Variability in {LLM}-Based Sentiment Analysis: Challenges, Mitigation Strategies, and the Role of Explainability},
  author={Herrera-Poyatos, David and Pel{\'a}ez-Gonz{\'a}lez, Carlos and Zuheros, Cristina and Herrera-Poyatos, Andr{\'e}s and Tejedor, Virilo and Herrera, Francisco and Montes, Rosana},
  journal={Frontiers in Artificial Intelligence},
  volume={8},
  pages={1609097},
  year={2025}
}

@article{wang2025materials,
  title={Evaluating the Performance and Robustness of {LLMs} in Materials Science {Q\&A} and Property Predictions},
  author={Wang, Hongchen and Li, Kangming and Ramsay, Scott and Fehlis, Yvonne and Kim, Edward and Hattrick-Simpers, Jason},
  journal={Digital Discovery},
  year={2025}
}

@article{yu2025reasonrobust,
  title={Benchmarking Reasoning Robustness in Large Language Models},
  author={Yu, Tong and Jing, Yongcheng and Zhang, Xikun and Jiang, Wentao and Wu, Wenjie and Wang, Yingjie and Hu, Wenbin and Du, Bo and Tao, Dacheng},
  journal={arXiv preprint arXiv:2503.04550},
  year={2025}
}

@article{li2025structflow,
  title={{StructFlowBench}: A Structured Flow Benchmark for Multi-Turn Instruction Following},
  author={Li, Jinnan and Li, Jinzhe and Wang, Yue and Chang, Yi and Wu, Yuan},
  journal={arXiv preprint arXiv:2502.14494},
  year={2025}
}

@article{wang2024understandingux,
  title={Understanding User Experience in Large Language Model Interactions},
  author={Wang, Jiayin and Ma, Weizhi and Sun, Peijie and Zhang, Min and Nie, Jian-Yun},
  journal={arXiv preprint arXiv:2401.08329},
  year={2024}
}

@article{gan2024clarqllm,
  title={{CLARQ-LLM}: A Benchmark for Models Clarifying and Requesting Information in Task-Oriented Dialog},
  author={Gan, Yujian and Li, Changling and Xie, Jinxia and Wen, Luou and Purver, Matthew and Poesio, Massimo},
  journal={arXiv preprint arXiv:2409.06097},
  year={2024}
}

@article{yang2025whatpromptsdontsay,
  title={What Prompts Don't Say: Understanding and Managing Underspecification in {LLM} Prompts},
  author={Yang, Chenyang and Shi, Yike and Ma, Qianou and Liu, Michael Xieyang and K{\"a}stner, Christian and Wu, Tongshuang},
  journal={arXiv preprint arXiv:2505.13360},
  year={2025}
}

@article{xu2024reliability,
  title={Reducing Tool Hallucination via Reliability Alignment},
  author={Xu, Hongshen and Zhu, Zichen and Pan, Lei and Wang, Zihan and Zhu, Su and Ma, Da and Cao, Ruisheng and Chen, Lu and Yu, Kai},
  journal={arXiv preprint arXiv:2412.04141},
  year={2024}
}

@inproceedings{zhang2025adversarial,
  title={From Allies to Adversaries: Manipulating {LLM} Tool-Calling through Adversarial Injection},
  author={Zhang, Ruian and Wang, Hao and Wang, Jiaxin and Li, Min and Huang, Yu and Wang, Dawei and Wang, Qi},
  booktitle={Proceedings of the 2025 Conference of the Nations of the Americas Chapter of the Association for Computational Linguistics: Human Language Technologies (NAACL-HLT)},
  pages={2009--2028},
  year={2025}
}

@article{zhu2025tooluse,
  title={Compounding Errors in Tool-Augmented Agents},
  author={Zhu, X. and others},
  journal={arXiv preprint},
  year={2025}
}

@article{song2024trialerror,
  title={Trial and Error: Exploration-Based Trajectory Optimization for {LLM} Agents},
  author={Song, Yifan and Yin, Da and Yue, Xiang and Huang, Jie and Li, Sujian and Lin, Bill Yuchen},
  journal={arXiv preprint arXiv:2403.02502},
  year={2024}
}

@inproceedings{wen2025scenarioindep,
  title={Scenario-Independent Uncertainty Estimation for {LLM}-Based Question Answering via Factor Analysis},
  author={Wen, Z. and Liu, Z. and Tian, Z. and Pan, S. and Huang, Z. and Li, D. and Huang, M.},
  booktitle={Proceedings of the ACM on Web Conference},
  pages={2378--2390},
  year={2025}
}

@article{lunardi2025reliability,
  title={On Robustness and Reliability of Benchmark-Based Evaluation of {LLMs}},
  author={Lunardi, Riccardo and Della Mea, Vincenzo and Mizzaro, Stefano and Roitero, Kevin},
  journal={arXiv preprint arXiv:2509.04013},
  year={2025}
}

@inproceedings{siska2024inadequacy,
  title={Examining the Robustness of {LLM} Evaluation to the Distributional Assumptions of Benchmarks},
  author={Siska, C. and Marazopoulou, K. and Ailem, M. and Bono, J.},
  booktitle={Proceedings of the 62nd Annual Meeting of the Association for Computational Linguistics (ACL)},
  pages={10406--10421},
  year={2024}
}

@article{wang2026agentnoisebench,
  title={{AgentNoiseBench}: Benchmarking Robustness of Tool-Using {LLM} Agents Under Noisy Condition},
  author={Wang, Ruipeng and Chen, Yuxin and Wang, Yukai and Wu, Chang and Fang, Junfeng and Cai, Xiaodong and Gu, Qi and Su, Hui and Zhang, An and Wang, Xiang and Cai, Xunliang and Chua, Tat-Seng},
  journal={arXiv preprint arXiv:2602.11348},
  year={2026}
}

@article{longcatflash2601,
  author       = {Meituan LongCat Team},
  title        = {LongCat-Flash-Thinking-2601 Technical Report},
  journal      = {CoRR},
  volume       = {abs/2601.16725},
  year         = {2026}
}

@article{yao2026coba,
  title={CoBA-RL: Capability-Oriented Budget Allocation for Reinforcement Learning in LLMs},
  author={Yao, Zhiyuan and Zhang, Yi-Kai and Chen, Yuxin and Sun, Yueqing and Xu, Zishan and Yang, Yu and Hu, Tianhao and Gu, Qi and Su, Hui and Cai, Xunliang},
  journal={arXiv preprint arXiv:2602.03048},
  year={2026}
}

@article{shi2025look,
  title={Look back to reason forward: Revisitable memory for long-context llm agents},
  author={Shi, Yaorui and Chen, Yuxin and Wang, Siyuan and Li, Sihang and Cai, Hengxing and Gu, Qi and Wang, Xiang and Zhang, An},
  journal={arXiv preprint arXiv:2509.23040},
  year={2025}
}

\clearpage
\appendix

\section{Training Configuration Details}
\label{appendix:training_config}

This section provides the complete training configurations for reproducing our experiments.

\subsection{Model and Infrastructure}

We use Qwen3-8B and Qwen3-32B as backbone models, both trained in BF16 precision with vLLM (v0.8.5) for efficient rollout generation.
All models use RoPE with $\theta=10^6$ and RMSNorm with $\epsilon=10^{-6}$.

\subsection{Optimization Hyperparameters}

Table~\ref{tab:optimization_config} summarizes the optimization hyperparameters shared across all methods.

\begin{table}[h]
\centering
\small
\caption{Optimization hyperparameters.}
\label{tab:optimization_config}
\begin{tabular}{lc}
\toprule
\textbf{Hyperparameter} & \textbf{Value} \\
\midrule
Optimizer & Adam \\
$\beta_1, \beta_2$ & $0.9, 0.95$ \\
Adam $\epsilon$ & $10^{-8}$ \\
Learning rate & $1 \times 10^{-6}$ \\
LR schedule & Constant \\
Weight decay & $0.01$ \\
Gradient clipping & $1.0$ \\
KL coefficient & $0.0$ \\
Discount factor $\gamma$ & $1.0$ \\
GAE $\lambda$ & $1.0$ \\
PPO epochs per step & 1 \\
Data reuse epochs & 2 \\
Gradient accumulation steps & 2 \\
Total training steps & 100 \\
\bottomrule
\end{tabular}
\end{table}

\subsection{Rollout and Generation Configuration}

Table~\ref{tab:rollout_config} details the rollout generation settings.

\begin{table}[h]
\centering
\small
\caption{Rollout and generation configuration.}
\label{tab:rollout_config}
\begin{tabular}{lc}
\toprule
\textbf{Parameter} & \textbf{Value} \\
\midrule
Training batch size & 16 \\
Rollouts per sample & 32 \\
Micro batch size & 1 \\
Max prompt length & 8,192 tokens \\
Max response length & 32,768 tokens \\
Max sequence length & 40,960 tokens \\
Sampling temperature (rollout) & 1.0 \\
Sampling temperature (eval) & 0.0 \\
Top-$p$ & 1.0 \\
Max interaction turns & 100 \\
\bottomrule
\end{tabular}
\end{table}

\subsection{Method-Specific Configurations}

Table~\ref{tab:method_config} compares the loss configurations across different training methods.

\begin{table}[h]
\centering
\small
\caption{Method-specific loss configurations.}
\label{tab:method_config}
\begin{tabular}{lccccc}
\toprule
\textbf{Method} & \textbf{Loss Type} & \textbf{Clip Range} & \textbf{Clip Ratio $c$} & \textbf{Adv. Norm} & \textbf{Loss Agg.} \\
\midrule
GRPO & grpo & $[0.2, 0.2]$ & 10.0 & batch & token-mean \\
DAPO & dapo & $[0.2, 0.28]$ & 3.0 & buffer & seq-mean-token-mean \\
GSPO & gspo & $[0.2, 0.28]$ & 3.0 & buffer & seq-mean-token-mean \\
Ours & gspo & $[0.2, 0.28]$ & 3.0 & buffer & seq-mean-token-mean \\
\bottomrule
\end{tabular}
\end{table}

For GRPO, we follow the original formulation with fixed clip range and batch-level advantage normalization.
DAPO and GSPO employ an asymmetric clip range $[0.2, 0.28]$ with dynamic temperature scaling and buffer-level advantage normalization, following their respective original implementations.
Our method inherits the GSPO loss configuration and adds the noise-aware curriculum on top.

\subsection{Noise-Aware Training Configuration}
\label{appendix:curriculum_config}

Our noise-aware training strategy consists of two components: controlled injection and noise scheduling, corresponding to the two factors described in Section~\ref{subsec:noise training}.

\paragraph{Controlled Injection.}
This component controls the noise scale $\rho = N_{\text{noise}} / N$, i.e., the proportion of perturbed rollouts within each task's rollout group.
For each task, $N$ rollouts are generated in parallel, among which $N_{\text{noise}}$ rollouts are executed in noisy environments and the remaining $N - N_{\text{noise}}$ in clean environments.
In our experiments, we fix the maximum noise proportion at 50\% of total rollouts (i.e., $\rho \leq 0.5$).
Training starts with $\rho = 0$, and the noise proportion is increased by a fixed step size each time the model's performance plateaus, as determined by the scheduling mechanism below.

\paragraph{Noise Scheduling.}
This component controls the noise difficulty and determines when to increase the noise scale.
We measure the model's robustness via the performance gap $\Delta$ between clean and perturbed rollouts (as defined in Section~\ref{subsec:noise training}):
when $\Delta$ falls below a predefined threshold $\theta$, the model is considered to have adapted to the current noise level.
Upon adaptation, we increase both the noise difficulty (characterized by the frequency of tool-side perturbations and the severity of user-side anomalies).
This yields a progressive curriculum that gradually increases interaction complexity while maintaining training stability.

\paragraph{Noise Types.}
We define noise along two axes: user-side noise (\emph{ambiguous}, \emph{inconsistent}, \emph{redundant}, \emph{out-of-scope}) and tool-side noise (\emph{failures}, \emph{incomplete}, \emph{misleading}, \emph{redundant}), as detailed in Section~\ref{subsec:noise injection}.

\subsection{Training Data Configuration}

For the multi-domain training setup (Table~\ref{tab:main_results_noisy} and Table~\ref{tab:main_results_ideal}), we train on tasks from two benchmarks simultaneously:
\begin{itemize}
    \item \textbf{$\tau^2$-Bench}: Airline, Retail, and Telecom domains, each filtered to medium-to-low pass rate difficulty.
    \item \textbf{VitaBench}: Delivery, In-Store, and OTA domains.
\end{itemize}

All baseline methods (GRPO, DAPO, GSPO) are trained exclusively on clean environments without noise injection.
In our method, noisy trajectories are progressively introduced via the controlled injection mechanism described above.
Groups where all rollouts receive identical rewards (all-pass or all-fail) are filtered out to ensure meaningful gradient signal.

\subsection{Evaluation Protocol}

Throughout both training and final evaluation, we use GPT-4.1 as the user simulator and Claude-Sonnet-4.5 as the evaluator.
During training, we evaluate every 5 steps with 4 rollouts per task, assessing both the ideal (noise-free) and noisy settings.
For the final evaluation reported in our main results, each experiment is repeated 4 times.
We report Avg@4 (average score across 4 runs) and Pass@4 (fraction of tasks solved in at least one of the 4 runs).

\subsection{Computational Resources}
\label{app:compute_resource}

For Qwen3-8B training, we use 32 NVIDIA H800 GPUs.
For Qwen3-32B training, we use 64 NVIDIA H800 GPUs.
Each training run of 100 steps takes approximately 3--5 days depending on the model scale and domain complexity.

\section{Discussion}

\subsection{Case Study}
\label{app:case_study}

We analyze a representative example from the $\tau^2$-Bench Retail domain to illustrate how noise affects agent behavior.
In this task, a user requests the return of two gaming-related items (a mechanical keyboard and a gaming mouse) from two separate orders, with refunds issued to the original payment method.
During the interaction, the environment injects intermittent API failures (e.g., \texttt{Error 429}) and corrupted fields.

As shown in Figure~\ref{fig:case_study}, both the base model and \ours successfully complete the information-gathering phase, including identity verification, order retrieval, and item identification.
However, their behaviors diverge in the execution phase.
The base model fails to invoke the return API and instead shifts to unrelated recommendations, resulting in task failure.
In contrast, \ours directly executes the required return operations once sufficient information is obtained, completing both returns successfully.

This example highlights a key behavioral difference under noise: while both models are capable of correctly understanding user intent, only \ours reliably translates this understanding into effective action.
Under noisy conditions, the base model fails to maintain focus on the primary objective and does not complete the required API calls.
In contrast, \ours remains aligned with the task objective and executes the necessary actions without deviation.

This phenomenon generalizes beyond the illustrated example.
Among the 23 tasks in this domain where \ours succeeds but the base model fails under noise, 78\% exhibit the same pattern: the base model correctly gathers all required information but fails to execute the critical action.
This suggests that noise primarily affects the transition from understanding to action, rather than the understanding itself.

Overall, the results indicate that curriculum training improves the agent's ability to maintain goal-directed behavior and reliably execute actions under noisy conditions, even when intermediate observations are corrupted or inconsistent.
\begin{figure}[t]
\centering
\small
\setlength{\fboxsep}{6pt}
\fbox{\parbox{0.93\columnwidth}{
\textbf{Task:} User requests return of gaming items from orders \#W5490111 and \#W7387996.

\vspace{6pt}
\textbf{Base model (failed, reward = 0.0):}
\vspace{2pt}

\textit{[Turns 1--11]} Agent correctly verifies the user's identity, retrieves both orders, and identifies the gaming items. User confirms: \emph{``Yes, please return the Mechanical Keyboard and the Gaming Mouse.''}

\textit{[Turn 15--30]} Instead of calling the return API, the agent starts recommending desk lamps and discussing student discounts. The conversation ends \emph{without any return being processed.}

\vspace{6pt}
\textbf{Our model (success, reward = 1.0):}
\vspace{2pt}

\textit{[Turns 1--18]} Agent verifies identity, retrieves orders, and summarizes the items with refund details. User confirms the same request.

\textit{[Turn 22]} Agent immediately executes:
\vspace{1pt}

\texttt{> return\_delivered\_order\_items(\#W5490111, [keyboard])}

\texttt{> return\_delivered\_order\_items(\#W7387996, [mouse])}

\vspace{1pt}
Both returns processed successfully in a single turn.
}}
\caption{Case study from $\tau^2$-Bench Retail (noisy setting). Both agents complete the information-gathering phase correctly, but the base model fails to execute the final action after encountering API noise, while our model stays on task.}
\label{fig:case_study}
\end{figure}

\subsection{Code of Ethics}
\label{app:ethics}

This work complies with the NeurIPS Code of Ethics. Our research focuses on improving the robustness of LLM-based agents under realistic, imperfect environments. All data used in this work are either synthetically generated by large language models or constructed through controlled pipelines, without involving any real user data or sensitive personal information.
To ensure responsible development, we adopt a strict data construction and validation process. Synthetic environments, interaction patterns, and noise perturbations are carefully designed to reflect realistic scenarios while avoiding harmful, unsafe, or misleading content. All generated data are further reviewed and refined to ensure consistency, correctness, and safety.
In addition, our experiments are conducted in controlled simulation environments, and the proposed framework does not directly interact with real users or external systems. Therefore, no human subjects are involved, and no ethical risks related to data privacy or user consent arise in this work.

\subsection{Broader Impacts}
\label{app:broader}

This work aims to improve the robustness of LLM-based agents under noisy and imperfect environments, which has important implications for real-world deployment. By exposing agents to diverse interaction uncertainties during training, our approach can lead to more reliable and adaptive systems in applications such as customer service, recommendation, and task automation.
However, improving agent robustness may also introduce potential risks. More capable and adaptive agents could be misused in scenarios requiring manipulation or exploitation of uncertain environments. In addition, if deployed without proper safeguards, robust agents may still propagate biases or make incorrect decisions under ambiguous inputs, potentially leading to negative user experiences or unintended consequences.
To mitigate these risks, our work focuses on controlled training settings and emphasizes the importance of structured evaluation and validation. The proposed framework is designed to improve generalization and reduce over-reliance on brittle patterns, which may contribute to safer and more reliable deployment. We encourage future work to further investigate fairness, safety, and alignment aspects when applying robust agent training methods in real-world systems.

\subsection{Safeguards}
\label{app:safeguards}

We implement several safeguards to ensure responsible use and construction of data and models in this work.
First, all environments, interaction data, and noise perturbations are generated through a controlled synthesis pipeline. The generation process is designed to avoid unsafe or harmful content, and all synthesized data are subject to validation and refinement to ensure correctness and consistency.
Second, our framework does not rely on real user data. All interaction patterns and task environments are synthetic, eliminating risks related to privacy leakage or misuse of personal information.
Finally, we adopt a modular design for noise injection, allowing controlled manipulation of user-side and tool-side perturbations. This prevents the introduction of uncontrolled or unrealistic behaviors that could compromise evaluation validity.

% \newpage
% \input{chapters/9_checklist}
\end{document}